\documentclass[twoside,journal]{IEEEtran}
\usepackage{times}
\usepackage{colortbl}
\usepackage{arydshln}
\usepackage{multirow}
\usepackage{multicol}
\usepackage{color,soul}
\usepackage{array}
\usepackage{yhmath}
\usepackage{amsmath,amssymb,amsfonts}
\usepackage{graphicx,subfigure,epstopdf,epsfig}
\usepackage{multirow,booktabs}
\usepackage[
            hypertexnames=false, 
            pdfstartview=FitH,
            CJKbookmarks=true,
            bookmarksnumbered=true,
            bookmarksopen=true,
            colorlinks, 
            linkcolor=red,
            anchorcolor=green,
            citecolor=blue
            ]{hyperref}
\usepackage{bookmark}
\usepackage{algorithm}
\usepackage{algorithmic}
\DeclareMathOperator*{\argmax}{argmax}
\graphicspath{{/}}

\newcommand{\eref}[1]
{(\ref{#1})}

\usepackage{cite}

\ifCLASSINFOpdf
\else
\fi
\begin{document}
%
\title{Sequential Dual Deep Learning with \\Shape and Texture Features for Sketch Recognition}
%
%

\author{Qi~Jia,~\IEEEmembership{Member,~IEEE,}
        Meiyu~Yu,
        Xin~Fan,~\IEEEmembership{Member,~IEEE,}
        and~Haojie~Li~\IEEEmembership{Member,~IEEE,}
\thanks{Q. Jia, X. Fan, and H. Li are with the DUT-RU International School of Information and Software Engineering, Dalian University of Technology, China, and also with the Key Laboratory for Ubiquitous Network and Service Software of Liaoning Province.\protect}
\thanks{M. Yu is with the School of Software, Dalian University of Technology, China. E-mail: xin.fan@ieee.org}
\thanks{This work is partially supported by the Natural Science Foundation of China under grant Nos. 61402077, 61572096, 61272371, 51579035, 61632006 and 61572105.}}

\maketitle

\begin{abstract}
Recognizing freehand sketches with high arbitrariness is greatly challenging. Most existing methods either ignore the geometric characteristics or treat sketches as handwritten characters with fixed structural ordering. Consequently, they can hardly yield high recognition performance even though sophisticated learning techniques are employed. In this paper, we propose a sequential deep learning strategy that combines both shape and texture features. A coded shape descriptor is exploited to characterize the geometry of sketch strokes with high flexibility, while the outputs of constitutional neural networks (CNN) are taken as the abstract texture feature. We develop dual deep networks with memorable gated recurrent units (GRUs), and sequentially feed these two types of features into the dual networks, respectively. These dual networks enable the feature fusion by another gated recurrent unit (GRU), and thus accurately recognize sketches invariant to stroke ordering. The experiments on the TU-Berlin data set show that our method outperforms the average of human and state-of-the-art algorithms even when significant shape and appearance variations occur.
\end{abstract}


%
\IEEEpeerreviewmaketitle

\section{Introduction}\label{sec:introduction}
%
%
%
%
Our ancestors used sketches to record their lives in ancient times. Nowadays, sketches are still regarded as an effective communicative tool. In past decades, researchers have intensively explore sketch characteristics in various multimedia applications including sketch recognition~\cite{DBLP:journals/tog/EitzHA12, schneider2014sketch}, sketch-based image retrieval~\cite{DBLP:journals/tvcg/EitzHBA11, DBLP:journals/cviu/HuC13} and sketch-based 3D model retrieval\cite{DBLP:conf/cvpr/WangKL15}.
Nevertheless, it is an extremely challenging task to recognize sketches drawn by non-artists even for human being. Firstly, sketches are highly abstract. For example, we can only use simple sticks to represent limbs of a human body in a sketch. Secondly, sketches have numerous styles that present significantly different appearances due to their free-hand nature. For a cow sketch, one may draw its coarse contour, while another may provide detailed patterns inside the outer contour. Last but not least, sketches are lack of texture cues. Lines and curves are enough to compose a sketch, showing evident differences with natural images.Unfortunately, most existing approaches to sketch recognition employ traditional feature extraction and classification techniques designated for textural images.

Researchers typically bring the features working pretty well for natural images into sketch recognition. These features include the histogram of oriented gradient (HOG)~\cite{DBLP:conf/cvpr/DalalT05} and scale-invariant feature transform (SIFT)~\cite{DBLP:journals/ijcv/Lowe04}. Both features highly depend on gradients of rich textures that rarely present in sketches composed of lines or curves. Recently, Yu~\emph{et~al.} developed Sketch-A-Net (SAN) that learns sketch features via the deep convolutional neural network (DCNN) inspired by the success of DCNN on recognizing handwritten digits~\cite{Yu2016Sketch}. These learnt features outperform those handcrafted ones, e.g., HOG and SIFT. However, DCNN generates features from the convoluted responses on image intensities (textures), which neglects the distinct geometrical structures in sketches. Object shapes representing intrinsic geometries are also stable to illumination and color variations. Therefore, the recognition rates of these methods lower than what human achieve can be partially owing to the absence of shape features~\cite{DBLP:journals/tog/EitzHA12}.

Most existing methods also neglect the inherently sequential nature of sketches besides their intrinsic geometries. In a recent work~\cite{Sarvadevabhatla2016Enabling}, Sarvadevabhatla~\emph{et~al.} built a recurrent network to sequentially learn features for recognition, yielding great improvements on accuracy. It is worth noting that sequentially drawn sketches exhibit a much higher degree of intra-class variations in stroke order. The learning strategy in~\cite{Sarvadevabhatla2016Enabling} fails tackling these high variations on stroke ordering. It is still unresolved to combat these variations when we learn sequential features for sketch recognition.

In this study, we devise dual recurrent neural networks with respect to textural and shape characteristics of sketches, and sequentially learn (actually combine) both features via joint Bayes for recognition. For the first time, coded shape features are introduced in a recurrent manner in order to distinguish sketches with similar textures but different shapes. Additionally, we explore the sequential nature of stroke groups rather than individual strokes so that stroke ordering is able to provide more information for recognition while excluding the effects from its variations. We validate our dual learning on the largest hand-free sketch benchmark TU-Berlin \cite{DBLP:journals/tog/EitzHA12}. Our method outperforms the state-of-the-art over 7 percentage points on the recognition rate.

\begin{figure*}
\begin{center}
\includegraphics[height=6cm,width=15cm]{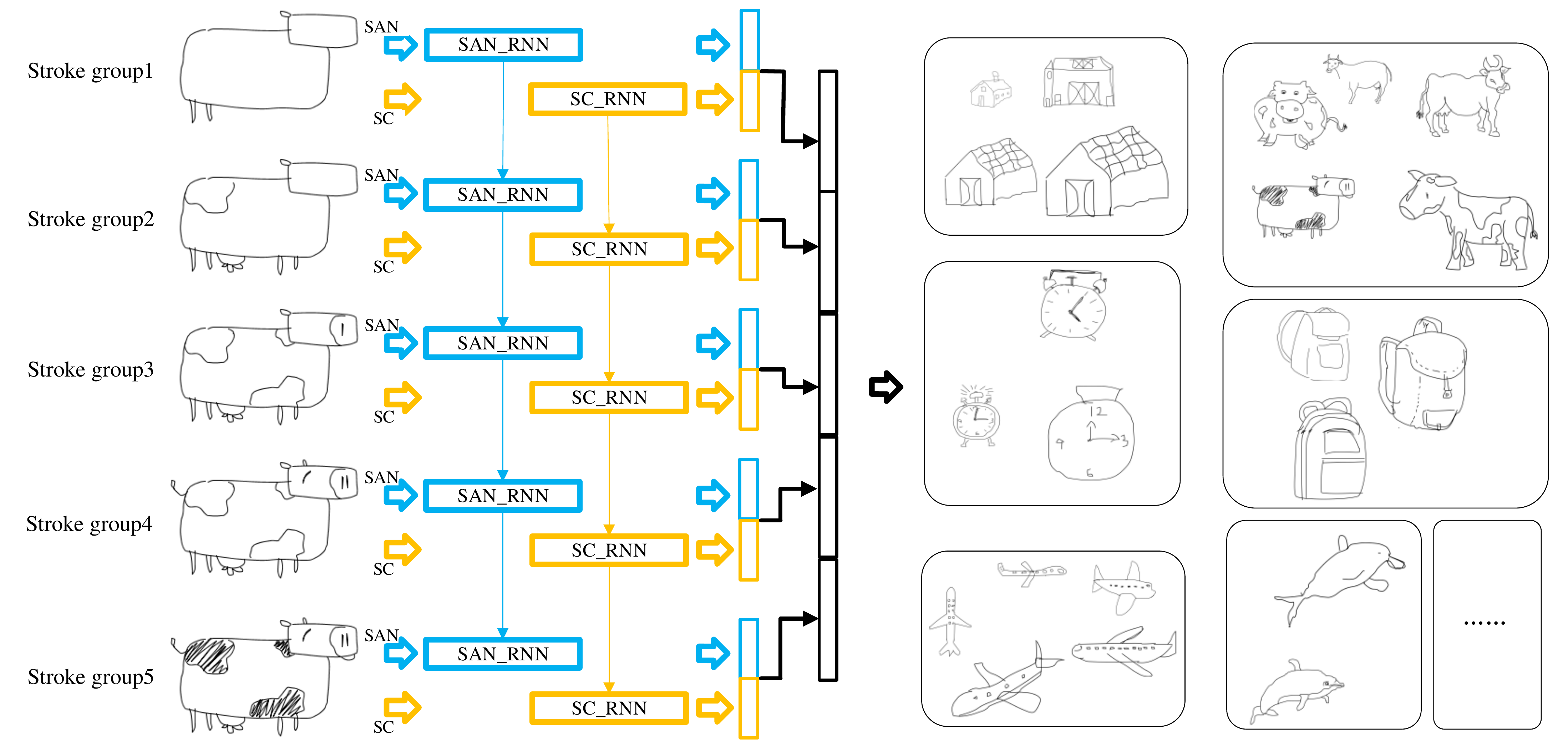}
\end{center}
\caption{Dual recurrent neural networks with respect to textural and shape features. These dual networks are sequentially trained by examples divided into five groups according to stroke ordering, and joint Bayes is applied for classification by the fused features from the networks.}

\label{fig:process}

\end{figure*}

The overall procedure is shown in  Fig.~\ref{fig:process}. Each sketch is divided into five groups according to its stroke ordering. We extract encoded shape context features~\cite{belongie2002shape} and texture features by Sketch-A-Net~\cite{DBLP:conf/bmvc/YuYSXH15} from each sketch group. Subsequently, dual recurrent neural networks upon gated recurrent units (GRUs) take textural and shape features as the inputs, respectively. The network receiving textural features is colored as blue in Fig.~\ref{fig:process}, while the one for shape features in yellow. And Then concatenate these two features by their time step. After this, we feed the concatenated features to another gated recurrent unit (GRU)(colored as blue and yellow). We sequentially train these networks group by group, and apply sumpooling to classify sketches upon the fused features.

\section{Related Work}
We review the features and deep learning architectures for sketch recognition.

\subsection{Features for sketch recognition}

In 2012, \cite{DBLP:journals/tog/EitzHA12} released a dataset containing 20,000 sketches distributed over 250 object categories. These sketches are drawn from daily objects, but humans can only correctly identify the sketch category with the accuracy 73\%. These results show that sketch recognition is a challenging task even for humans. Features for sketch recognition can be roughly categorized into handcrafted ones and those learned from deep learning.

Hand-crafted features share similar spirit with image classification methods, including feature extraction and classification. Most of the existing works regard sketch as texture image, such as HOG and SIFT~\cite{DBLP:conf/cvpr/DalalT05}~\cite{DBLP:journals/ijcv/Lowe04}. In~\cite{DBLP:journals/tog/EitzHA12}, SIFT feature are extracted in image patches. The method also takes into account that sketches do not have smooth gradients and are much sparser than images. Similarly,~\cite{schneider2014sketch} leverage Fisher Vectors and spatial pyramid pooling to represent SIFT features. \cite{DBLP:journals/cviu/LiHSG15} employ multiple-kernel learning (MKL) to learn appropriate weights of different features. The method improve the performance a lot as different features are complementary to each other.

Recently, deep neural networks (DNNs) have achieved great success\cite{DBLP:conf/nips/KrizhevskySH12}by replacing hand-crafted representation with learning strategy.\cite{DBLP:conf/bmvc/YuYSXH15} ~\cite{Yu2016Sketch} leverage a well-designed convolutional neural network(CNN) architecture for sketch recognition.According to their experimental results, their method surpass the best result achieved by human. According to their experimental results, their method surpass the best result achieved by human.In~\cite{johnson2009computational} ~\cite{schneider2014sketch} , the sequential information of sketch is exploited.. \cite{Sarvadevabhatla2016Enabling} explicitly uses sequential regularities in strokes with the help of recurrent neural work (RNN).

The experimental results show that they get the state-of-art performance on sketch recognition task. However, all the methods upon deep learning only rely on texture information. The convolution process may confuse sketches with similar texture but different contours. The variation on stroke orders of  the same class also affect the convergence of RNN.

\subsection{Learning architectures}

CNNs and RNNs are two branches of neural network. CNNs have obtained great success in many practical application~\cite{DBLP:conf/nips/KrizhevskySH12,DBLP:journals/corr/SimonyanZ14a}. LeNet~\cite{DBLP:conf/nips/CunBDHHHJ89} is a classical CNN architecture, which has been used in handwritten number recognition. Wang \emph{et al.}~\cite{DBLP:conf/cvpr/WangKL15} use Siamese network to retrieve 3D model by sketch. However, all the above methods take sketches as traditional image and ignore the sparse of sketches.

In order to exploit appropriate learning architecture for sketch recognition, \cite{DBLP:conf/bmvc/YuYSXH15}~\cite{Yu2016Sketch} develop a network named as Sketch-A-Net (SAN), which enlarges the pooling sizes and patches of filters to cater the sparse character of sketches. Different from traditional images, sketches involve inherent sequential property. \cite{DBLP:conf/bmvc/YuYSXH15} and \cite{Yu2016Sketch} divide the strokes in several groups according to the strokes order. However, CNNs can not build connections between sequential strokes.

The sequential property is not proprietary for sketches. Many works resort to RNN to improve the performance in speech recognition~\cite{vinyals2012revisiting}and text generation~\cite{DBLP:conf/icml/SutskeverMH11}. RNN is specialized for processing input sequence, which can bridge the hidden units and deliver the outputs from former sequence to the latter. However, it has a significant limitation called 'vanishing gradient'. When the input sequence is quite long, RNN is difficult to propagate the gradients through deep layers of the neural network, which is easy to cause gradients vanishing and exploding problems~\cite{Graves1997Long}. In order to overcome the limitation of RNN, long short term memory (LSTM)~\cite{Graves1997Long} and gated recurrent unit (GRU)~\cite{cho2014learning} are proposed. GRU can be regarded as a light-weight version of LSTM, which outperforms LSTM in some certain cases by learning smaller number of parameters~\cite{chung2014empirical}.

Sarvadevabhatla~\emph{et al.}~\cite{Sarvadevabhatla2016Enabling} take orders of strokes as a sequence and feed their features to GRU. In this way, a long-term sequential and structural regularities of stoke can be exploited. As a result, they achieve the best recognition performance. However, stroke order varies severely in the same kind, which results in the fluctuation of the network.

\begin{figure*}
\begin{center}
\includegraphics[height=6cm,width=18cm]{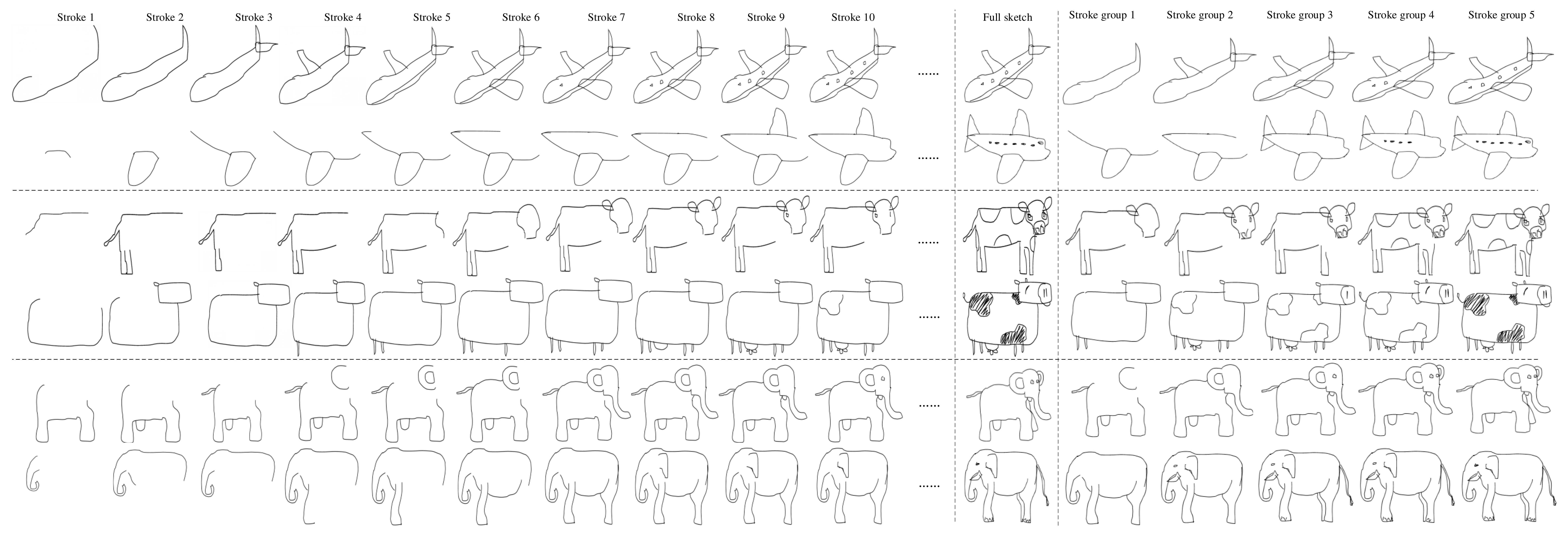}
\end{center}
\caption{Comparisons of stroke by stroke and stroke groups}

\label{fig:stroke_group}
\end{figure*}

\section{sequential learning framework}

In this section, we introduce the architecture of our dual sequential learning framework. We first illustrate the recurrent neural network based on Gated Recurrent Unit (GRU).Then, we exploit the stroke order and describe strokes by texture and shape features. Finally, the coupled features are combined another (GRU).

\subsection{Gated Recurrent Unit}

In our work, we take GRU as basic network architecture and then exploit the sequential characteristic of the sketches.A GRU network learns how to map input sequence ${X_T}=\left( {{x_1},{x_2}, \ldots ,{x_T}} \right)$ to output sequence ${Y_T}=\left( {{y_1},{y_2}, \ldots ,{y_T}} \right)$. This mapping is illustrated in ~\eref{eq:gru1} and  ~\eref{eq:gru5}.
\begin{equation}\label{eq:gru1}
    {r_t} = sigm\left( {{W_{xr}}{x_t} + {W_{hr}}{h_{t - 1}} + {b_r}} \right),
\end{equation}
\begin{equation}\label{eq:gru2}
    {z_t} = sigm\left( {{W_{xz}}{x_t} + {W_{hz}}{h_{t - 1}} + {b_z}} \right),
\end{equation}
\begin{equation}\label{eq:gru3}
    \mathop {{h_t}}\limits^ \sim   = \tanh ({W_{xh}}{x_t} + U\left( {{r_t} \odot {h_{t - 1}}} \right) + {b_h}),
\end{equation}
\begin{equation}\label{eq:gru4}
    {h_t} = \left( {1 - {z_t}} \right) \odot {h_{t - 1}} + {z_t} \odot \mathop {{h_t}}\limits^ \sim,
\end{equation}
\begin{equation}\label{eq:gru5}
    {y_t} = {W_{hy}}{h_t},
\end{equation}

where $x_t$ is the $t\emph{th}$ input and $y_t$ is the $t\emph{th}$ output. $h_t$ is the 'hidden' state of GRU and regulated by gating units $r$, $z$ and $\mathop {{h_t}}\limits^ \sim$. The operation $\odot$ denotes the element-wise vector product. $W_*$, $U$ are weight matrices and $b_*$ is the weight vector for GRU. More details about GRU can be found in~\cite{chung2015gated}.

\subsection{Sequential dual learning architecture based on GRU}
In this section, the strokes are first divided into several groups to reduce the effects on the variation of strokes in time sequence. Then, two features of texture and coded shape are introduced to characterize the texture and geometry information of sketches.
\begin{figure*}
\centering
\includegraphics{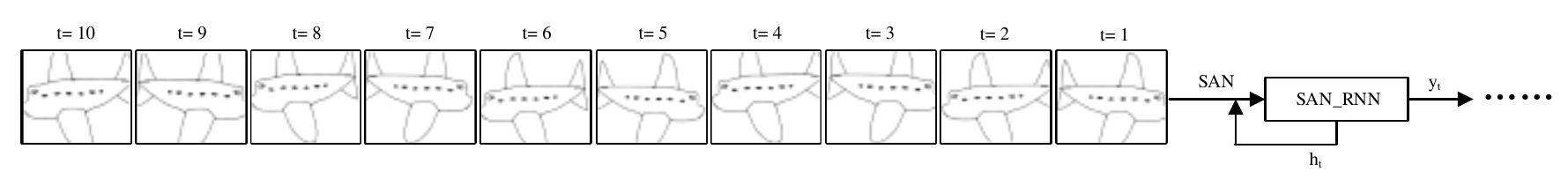}
\caption{Input sequence of one stroke group.}
\label{fig:partto10}
\end{figure*}

\subsubsection{Stroke order exploitation}

For traditional images, all the pixels are captured at the same time. However, sketch consists of sequential strokes and the order of strokes is an important information for sketch. When drawing the same object, different people have their own orders of strokes~\cite{DBLP:journals/tog/EitzHA12}. As shown on the left of Fig.~\ref{fig:stroke_group}, there are 3 pairs of sketches: two airplanes, two cows and two elephants. The figure is divided into three parts and they are divided by 2 vertical dash lines. The middle part is the complete sketches. The left part shows the first ten strokes of the sketches. We can see that the same objects have different stroke orders. If each stroke is taken as the input of GRU directly, the difference of the input may make the network unstable. The right part shows the stroke groups which are based on stroke orders, the first column shows the first 20\% strokes, and the second column shows the first 40\% in stroke sequence. And so on, the last column shows the complete sketches. We can see the difference of the sketches in the same time step is reduced.

In order to reduce the effect of stroke order, each sketch is divided into five sketches according to the time sequence. Suppose there are $N$ strokes for sketch $S$, which can be represented as $\left( {{s_1},{s_2}, \ldots ,{s_{N - 1}},{s_N}} \right)$. Then, the first extended sketch group contains strokes from $s_1$ to $s_{N/5}$, the second sketch contains strokes from $s_1$ to $s_{2N/5}$, accordingly, the last one contains the whole sketch of the original, as shown in Fig.~\ref{fig:process}. For robustness, we also take 10 crops and reflections for each train and test sketches~\cite{krizhevsky2012imagenet}. After this operation, each sketch is derived into 10 sketches in each group. Hence, we have 50 sketches in total, represented as $S = \left( {s{s_1},s{s_2}, \ldots ,s{s_{49}},s{s_{50}}} \right)$, and each input sequence contains 10 sketches. Figure.~\ref{fig:partto10} shows an input sequence of one stroke group. From right to left, when $t$ is odd, the inputs are crops of original stroke group and when $t$ is even, the inputs are crops of original stroke group's reflection. The order for each is top left, bottom left, top right, bottom right and center.

\subsubsection{Texture and shape features}

Sketches encode both texture and geometry information. The texture information is used to characterize details of sketches, while shape information is used to obtain geometry feature. Both are necessary for sketch recognition. For example, basketball and football share similar out contour but different texture inside, while parrots and pigeon have similar wings but different beaks. Hence, two features are mutual complementary.
\begin{figure}[t]
	\begin{center}
		\includegraphics[width=0.8\linewidth]{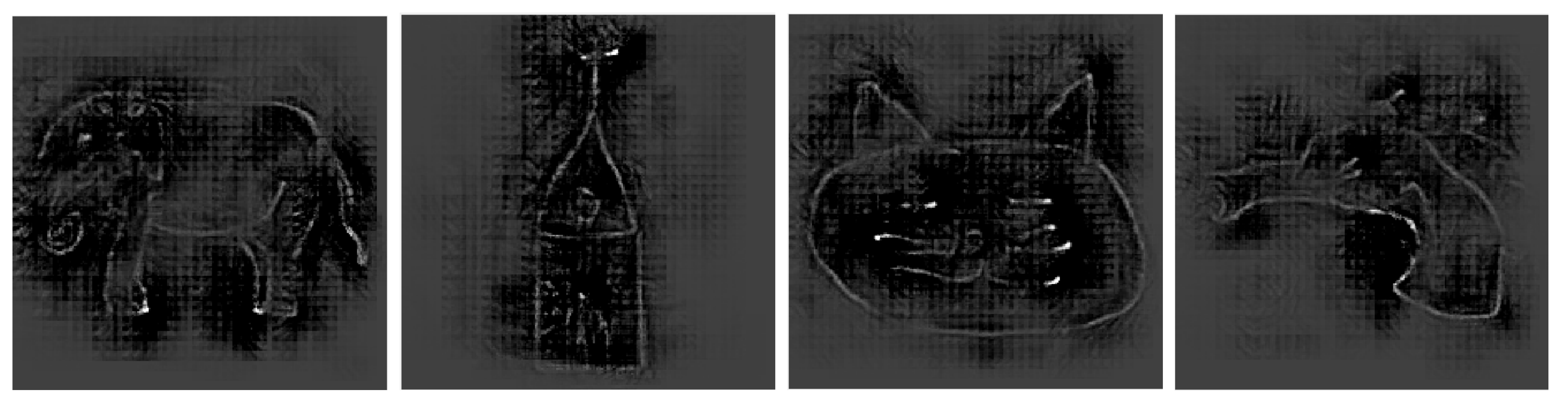}
	\end{center}
	\caption{Visualisation of the learned filters by deconvolutional.}
	\label{fig:deconvolutional}
\end{figure}

\begin{figure}[t]
	\begin{center}
		\includegraphics[width=0.8\linewidth]{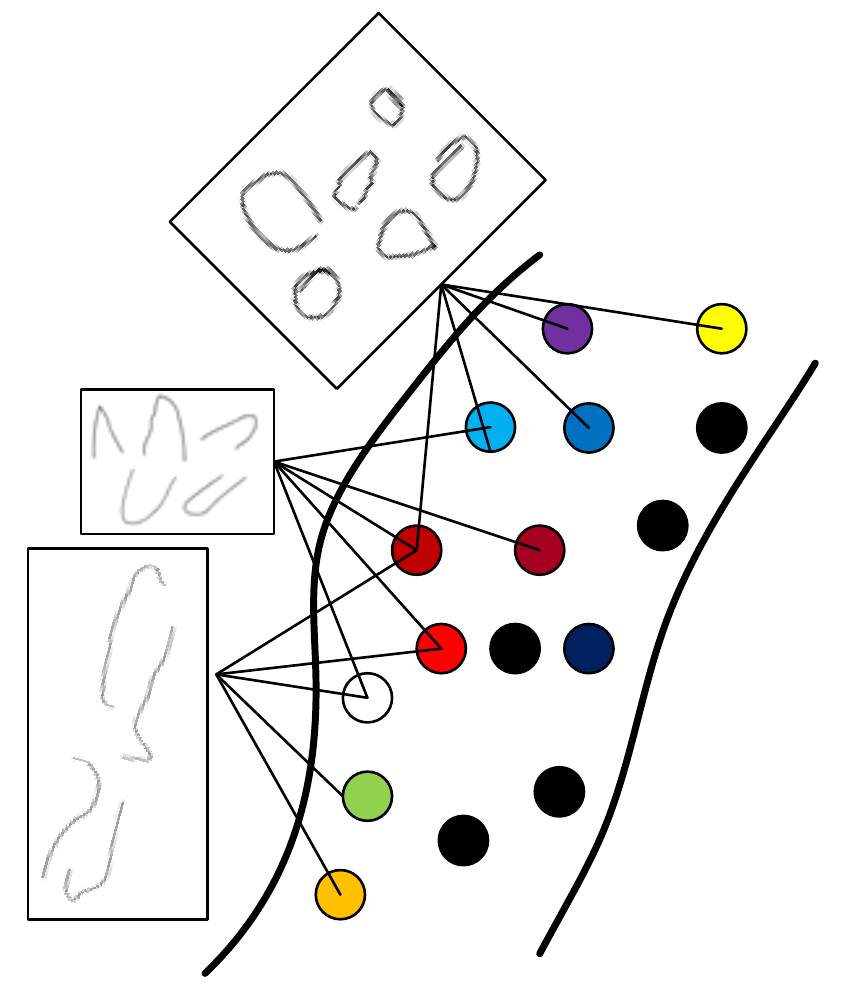}
	\end{center}
	\caption{Encoded features of similar shapes.}
	\label{fig:codebook}
\end{figure}

We use the newly developed Sketch-A-Net (SAN)~\cite{DBLP:conf/bmvc/YuYSXH15} to obtain texture feature, which is based on CNN architecture. And the shape feature is represented by coded shape context~\cite{belongie2002shape}, which makes our method robust to stroke variations. Each kind of features are taken as input of two GRUs respectively.

\textbf{Texture feature} is used to represent texture information of image, which is an important feature for recognition. Recently, texture features extracted by CNNs provide strong discrimination on sketch recognition. Thus, in this paper, we use Sketch-A-Net~\cite{DBLP:conf/bmvc/YuYSXH15} to extract texture features of sketch. Sketch-A-Net has 8 layers, the first five layers are convolutional layers and the last three layers are fully connected layers, each convolutional layer with rectifier (ReLU) units, while the first, the second and the fifth layers are followed by max pooling. The final layer has 250 output units, which are corresponding to the number of unique classes in the TU-Berlin~\cite{DBLP:journals/tog/EitzHA12} dataset.

In order to get texture features of sketch, we feed each sketch in sequence into Sketch-A-Net and take the 512 dimensional features extracted from the last fully-connected layer to represent texture features of sketch. The texture features of the sketch are denoted as $san$. In order to observe what has been learned by Sketch-A-Net, we make a deconvolution on the filters of the fifth layer. As shown in Fig. ~\ref{fig:deconvolutional}, the filter can capture complex texture features of object.

\textbf{Shape feature} is used to describe geometry information of sketches. As the high flexibility of sketches, each stroke can be drawn in different styles. As shown in Fig.~\ref{fig:codebook}, strokes in each rectangle are the same kind with different variations. In order to handle intra-class variation of strokes, we leverage a coding method~\cite{wang2014bag} to make similar strokes represented by similar features.

First of all, shape context~\cite{belongie2002shape} is employed as the geometry metric for strokes. Each stroke is represented by shape context features with dimension $d$. Then, the feature of $i$\emph{th} stroke can be represented as $s{t_i} \in {R^{d \times 1}}$$\left( {i = 1,2 \ldots ,n} \right)$, and $n$ is the number of the stroke.

Secondly, we use k-means~\cite{duda2000pattern} to obtain codebook based on low-level geometry features. We apply the k-means algorithm on randomly selected shape features. The cluster centers are regarded as the codebook, as shown in Equ.~\eqref{codebook}. $b_j({j = 1,2 \ldots ,M})$ is denoted as the clustering center and there are $M$ cluster centers in total. Thereafter, we can use $M$ prototypes to describe the whole stroke space. As shown in Fig.~\ref{fig:codebook}, the colorful circle points stand for the cluster centers, and each sketch can be represented by several of them.
\begin{equation}\label{codebook}
B = \left( {{b_1},{b_2}, \ldots ,{b_M}} \right) \in {R^{d \times M}},
\end{equation}

In order to get the final representation of strokes, we use codebook to encode shape features. By comparing two classical coding method vector quantization coding (VQ)~\cite{lazebnik2006beyond} and Local-constraint linear coding (LLC)~\cite{wang2010locality}, we choose the latter one, which is more fast and effective.

We uses $k$ nearest neighbors of $st_i$ in codebook $B$ (in Fig.~\ref{fig:codebook}, $k$ is 5) to reconstruct $st_i$, which can be denoted as $\mathop {NN}\nolimits_{\mathop {ind}\nolimits_i }  \in {R^{d \times k}} $. Here $\mathop {ind}\nolimits_i  = \left( {\mathop {ind}\nolimits_i^1  \cdots \mathop {ind}\nolimits_i^k } \right)$ is a set which contains $k$ nearest neighbors in $B$. While $\mathop {NN}\nolimits_{\mathop {ind}\nolimits_i }$ is a matrix, which consists of ${\mathop {ind}\nolimits_i^1  \cdots \mathop {ind}\nolimits_i^k }$ columns of $B$. Furthermore, we can obtain the linear coefficients $\mathop w\nolimits_{\mathop {ind}\nolimits_i }  \in {R^{k \times 1}}$ of columns in $\mathop {NN}\nolimits_{\mathop {ind}\nolimits_i }$ by optimizing Equ.~\eqref{targetfucntioninLLC}. As shown in Fig.~\ref{fig:codebook}, similar strokes may have the same $k$ nearest neighbors.

 \begin{equation}\label{targetfucntioninLLC}
 \mathop {\min }\limits_{\mathop {ind}\nolimits_i } \mathop {\left\| {\mathop {st}\nolimits_i  - \mathop {NN}\nolimits_{\mathop {ind}\nolimits_i } \mathop w\nolimits_{\mathop {ind}\nolimits_i } } \right\|}\nolimits^2 \quad s.t.\quad \mathop 1\nolimits^{\rm T} \mathop w\nolimits_{\mathop {ind}\nolimits_i }  = 1
 \end{equation}

Finally, we get more discriminative features by combining LLC and max-pooling~\cite{huang2014feature}. In our work, we run max-pooling on all stroke features and the final shape feature can be represented as $sc$.For each sketch ${ss}_t$($t \ge 1$) in sketch sequence, we can extract both texture and shape features denoted as ${san}_t$ and ${sc}_t$, respectively.

\subsection{Dual Deep Learning Strategy}
As texture and shape feature are mutual complementary to each other, we leverage a dual deep learning strategy to combine two features. Specifically, we use two separated RNNs to take each feature as input, and combine them at output by time-based weights.
In each time step $t$ ($ 1\le t \le 50$), both texture and shape RNNs can generate features to represent input sketches, denoted as $\mathop {fsan}\nolimits_t$ and $\mathop {fsc}\nolimits_t$, respectively. In order to utilize texture and shape features, we concatenate these two generated features by their timesteps. And then we feed concatenated features to another GRU ($SANSC$)to learn relevance of different features and different timesteps.

After processed by SANSC, every timestep$t$ ($1 \le t \le 50$) can give a prediction vector, denoted as $y_t$ of the input sketch. We sum over all prediction vectors denoted as $S$(see equation \eqref{sumofprediction})and choose the class $pc$ with max value in $S$(see equation \eqref{maxindex}) as prediction
\begin{equation}\label{sumofprediction}
{S} = \sum\limits_{t=1}^{50}{y_t}
\end{equation}
\begin{equation}\label{maxindex}
{pc} = \mathop{\argmax}_{pc}{S_[pc]}
\end{equation}
As texture and shape feature are mutual complementary to each other,we leverage a dual deep learning strategy to combine two features. As texture and shape feature are mutual complementary to each other, we leverage a dual deep learning strategy to combine two features. Specifically, we use two separated RNNs to take each feature as input, and combine them at output by time-based weights.

\section{Experiments and Results}
In this section we first provide the dataset and experiment setting of our method. Then, our method is compared with the state-of-the-art based on the same protocol. Finally, we validate the effectiveness of the shape feature
The results show our method outperform the other methods at 7\% in recognition rate.

\subsection{Dataset and data augmentation}

We evaluate our method on the TU-Berlin sketch dataset~\cite{DBLP:journals/tog/EitzHA12}, which is the most largest and commonly used human sketch dataset. It consists of 20000 sketches and 250 categories (80 sketches per category). The dataset was collected on Amazon Mechanical Turk(AMT) from 1350 participants. Thus the dataset guarantees the diversity of object categories and sketching styles within every category. We use $67\%$ of data for training, $13\%$ for evaluation and the rest of sketches are used for testing.

In our experiments, data augmentation is employed to reduce overfitting. In order to increase the number of sketches per category, we apply several transformations on each sketch, including horizontal reflection and rotation ([-5,-3,0,+3,+5] degrees). Then we do systematic combinations of horizontal and vertical shifts ($\pm 15$ pixels). The data augmentation procedure results in $18\left( {10 + 8 = 18} \right) \times 80 = 1440$ sketches per category, a total number of $1440 \times 250 = 360000$ sketches distributed over 250 categories.

\subsection{Experiment settings}
In our method, we use Sketch-A-Net~\cite{DBLP:conf/bmvc/YuYSXH15} framework to extract texture features of sketches. The framework ~\cite{DBLP:conf/bmvc/YuYSXH15} consists of five different CNNs which are independently trained for five different scaled versions of original sketch. In our experiment, we employ the single channel network which is trained for sketches with $256 \times 256$ pixels  and we take 512-dimensional features of the last fully-connected layer to represent sketches. For shape features, we apply shape context on $5$ points with equal interval of each stroke. The number of bins of shape context is set to be $60$. Thus, the dimension of shape context descriptor for each stroke is $300$. The size of codebook is $500$. We implement our network based on Torch\cite{collobert2002torch}. The initial learning rate is set to be $0.002$ and batch size is $100$.

\subsection{Comparison with the state-of-the-art}

We compare our method with the state-of-the-arts, which are used for sketch recognition. These methods can be divided into two groups, one group combines hand-crafted features and classifiers, including HOG-SVM method~\cite{DBLP:journals/tog/EitzHA12}, structured ensemble matching~\cite{li2013sketch}, multi-kernel SVM~\cite{DBLP:journals/cviu/LiHSG15} and Fisher Vector Spatial Pooling (FV-SP)~\cite{schneider2014sketch}. The other group is DNN-based methods, including AlexNet~\cite{DBLP:conf/nips/KrizhevskySH12}, LeNet~\cite{DBLP:conf/nips/CunBDHHHJ89}, AlexNet-FC-GRU~\cite{DBLP:conf/mm/Sarvadevabhatla16b}, and two versions (SN1.0~\cite{DBLP:conf/bmvc/YuYSXH15} and SN2.0~\cite{Yu2016Sketch}) of Sketch-A-Net framework. The experiments results of these methods are obtained by their executive implements or the reported results in their papers.

\begin{table}
\centering
 \caption{Accuracy of Sketch Recognition on TU-Berlin Dataset}
  \label{tab:recognitionaccuracyrate}
  \begin{tabular}{ccl}
    \toprule
    Method & Recognition Accuracy(\%) \\
    \midrule
    HOG-SVM~\cite{DBLP:journals/tog/EitzHA12} & 56.0 \\
Ensemble~\cite{li2013sketch} & 61.5 \\
MKL-SVM~\cite{DBLP:journals/cviu/LiHSG15} & 65.8 \\
FV-SP~\cite{schneider2014sketch} & 68.9 \\
Humans~\cite{DBLP:journals/tog/EitzHA12} & 73.1 \\
AlexNet-SVM~\cite{DBLP:conf/nips/KrizhevskySH12} & 67.1 \\
AlexNet-Sketch~\cite{DBLP:conf/nips/KrizhevskySH12} & 68.6 \\
LeNet~\cite{DBLP:conf/nips/CunBDHHHJ89} & 55.2 \\
SN1.0~\cite{DBLP:conf/bmvc/YuYSXH15} & 74.9 \\
SN2.0~\cite{Yu2016Sketch} & 78.0 \\
AlexNet-FC-GRU~\cite{DBLP:conf/mm/Sarvadevabhatla16b} & 85.1  \\
Our Method & 92.2 \\
  \bottomrule
\end{tabular}
\end{table}

Table~\ref{tab:recognitionaccuracyrate} shows the recognition accuracy of the comparable methods on TU-Berlin dataset. In general, methods based on DNN gain better performance than the ones based on hand-crafted features. The accuracy of the methods based on hand-crafted features is 63\% in average, which is lower than the results of Humans~\cite{DBLP:journals/tog/EitzHA12}. This is because the existing hand-crafted features are well-designed for traditional images but not fit for abstract and sparse sketches. In contrast, the accuracy in average for DNN based methods is about 74\%, which is about 1\% higher than Humans~\cite{DBLP:journals/tog/EitzHA12}, and 11\% higher than the feature based group. Among the existing DNN based methods, SN1.0 is the first method that beats humans with the accuracy 74.9\%, and the improved version (SN2.0) obtain the accuracy 77.95\%. As far as we know,~\cite{DBLP:conf/mm/Sarvadevabhatla16b} is the sate-of-art method with recognition accuracy of 85.1\%.

Our method shares the advantages of DNN based methods, and outperform the other methods with classification accuracy 92.2\%, which is about 7\% higher than the state-of-the-art ~\cite{DBLP:conf/mm/Sarvadevabhatla16b}. Thus, the results illustrate the effectiveness of our dual learning strategy and the coupled features.

\subsection{Effects of shape features}
\par
\begin{figure}[t]
	\begin{center}
		\includegraphics[width=0.8\linewidth]{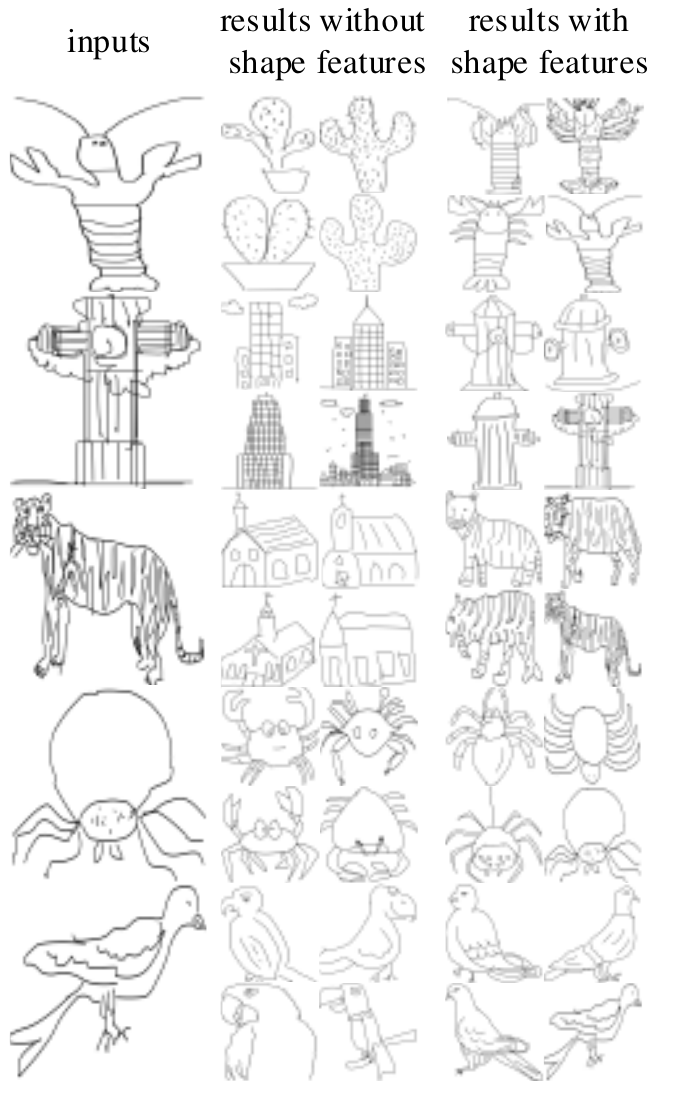}
	\end{center}
	\caption{Classification results of method without shape features and our method }
	\label{fig:wholeshape}
\end{figure}

In order to validate the effect of the shape features in sketch recognition, we list the results of our architecture with and without shape features in Tab.~\ref{tab:contributionsofshapefeatures}. We can see that the accuracy without shape features is only 84.1\%, which is about 7\% lower than the one with shape features. This indicates shape feature plays an important role in sketch recognition.
\begin{table}
    \centering
	\caption{Evaluation on the Contributions of Shape Features}
	\label{tab:contributionsofshapefeatures}
	\begin{tabular}{ccl}
		\toprule
		Method & Recognition Accuracy (\%) \\
		\midrule
		Without Shape Features & 84.1 \\
		With Shape Features & 92.2 \\
		\bottomrule
	\end{tabular}
\end{table}
In order to make the effects of shape feature more intuitive, we also exhibit some samples of the recognition results without and with shape features in Fig.~\ref{fig:wholeshape}. The first column lists the query sketches, including lobster, fire hydrant, tiger, spider, and pigeon. The second column shows the classification results without shape features and the last column is the results of our method with shape features. Each class are represented by four sketches in it. We can see that the method without shape features produces wrong classification results for all the requests. The reason is texture features have low distinctive ability to sketches with similar textures, such as horizontal and vertical lines on fire hydrant and skyscraper and stripe on tiger and church. In contrast, the method with shape features obtain correct classifications for all requests, and it is more distinctive than the methods without shape features. Our method can even distinguish the vivid difference at the beak of pigeon and parrot.
Method without shape features gives the wrong results and there is an important connection between input sketch and results of method without shape features, that is, they have similar texture.

\begin{figure}[t]
	\begin{center}
		\includegraphics[width=0.8\linewidth]{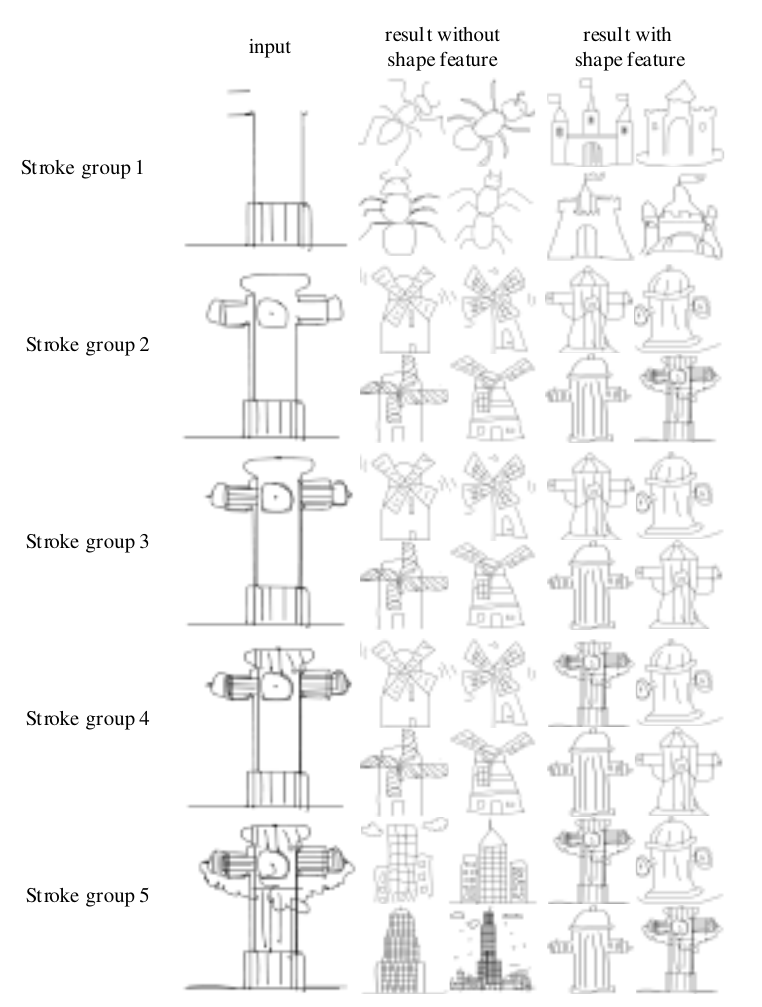}
     \end{center}
	\caption{Classification results of method without shape features and our method in different time steps. First column:input, second column : results without shape features, the last column: results with shape features}

	\label{fig:timestep}
\end{figure}

Further, we illustrate the classification results of each outputs. As shown in Fig.~\ref{fig:timestep}, the first column shows the query sketches represented by five groups, the second and the third columns show the results for each outputs without and with shape features. We can see that the results of each output in the second column are wrong, while ours produce correct results from the second output of the time sequence. The methods in this experiment are implemented without JB, which can verify that the shape feature can work effectively without the help of JB.

\begin{figure}[t]
	\begin{center}
		\includegraphics[width=0.8\linewidth]{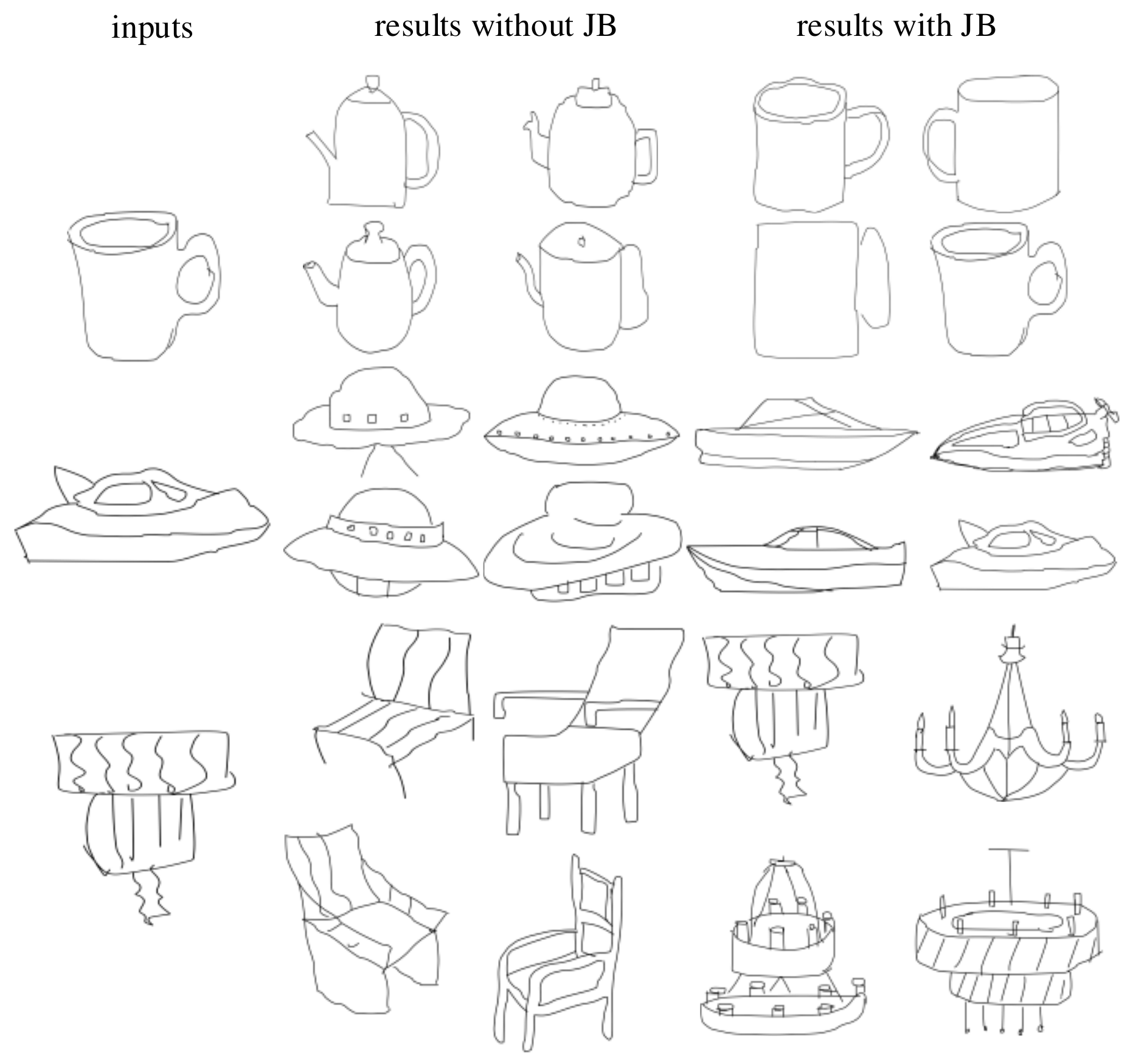}
	\end{center}
	\caption{Classification results of method without Joint Bayesian and with Joint Bayesian.}
	\label{fig:results_with_withoutJB}
\end{figure}

\subsection{Effects of Joint Bayesian}

In this section, the effects of Joint Bayesian is illustrated. As shown in Tab.~\ref{tab:contributionsofJB}, the results with and without JB are listed separately. The recognition result of our method without JB is achieved by softmax layer. The recognition rate is 88.2\%, which is still higher than~\cite{DBLP:conf/mm/Sarvadevabhatla16b}. The results indicate that JB make features from different time steps have different weights. With the help of JB, features from different time steps are fused together. JB can provide the weights between different features, and assignment important features with higher weights. As shown in Fig. ~\ref{fig:results_with_withoutJB}, we choose some recognition results of method without JB and with JB. The sketches in the first column represent query sketches, they are mug, speed-boat and chandelier. The sketches in the second column and third column represent samples of classification category given by two methods respectively. We can see that classification results given by method with JB are all right, however method without JB gives the wrong answers. The reason is that classification results are influenced by texture and shape features from all time steps of one sketch. Different features should have different contributions to classification. However, method without JB could not give the contributions of all features and is easy to give the wrong classification results. Such as in Fig. ~\ref{fig:results_with_withoutJB}, mug and teapot have similar shapes and textures, speed-boat and flying saucer also have similar shaped and features, method without JB could not leverage the weights of all features. And as for chandelier and armchair, they have the same texture, that is vertical lines, shape features are disadvantaged to determine classification. Compared to method without JB,  method with JB can learn weights of all features and give the right classification.

\begin{table}
  \centering
  \caption{Evaluation on the contributions of JB}
  \label{tab:contributionsofJB}
  \begin{tabular}{ccl}
    \toprule
    Method & Recognition Accuracy (\%)\\

\midrule
Without JB & 88.2 \\
With JB & 92.2 \\
  \bottomrule
\end{tabular}
\end{table}

\section{Conclusion}
In this paper, we proposed a sequential dual deep learning strategy combined both shape and texture features for sketch recognition. According to our experimental results, we achieve the best performance on sketch recognition. Our method has the following advantages. First, we employ texture feature and shape feature to characterize texture and geometry information of one sketch, which are complementary to each other. Second, We explore the sequential nature of sequential stroke groups rather than sequential strokes. The learned features of sketch can be used in some other sketch-related applications, such as sketch-based image retrieval and 3D shape retrieval.


%





\ifCLASSOPTIONcaptionsoff
  \newpage
\fi



\bibliographystyle{IEEEtran}
\bibliography{sketch}
%

%








\end{document}